\pdfoutput=1

\documentclass[11pt]{article}

\usepackage[final]{acl}

\usepackage{times}
\usepackage{latexsym}

\usepackage[T1]{fontenc}

\usepackage[utf8]{inputenc}

\usepackage{microtype}

\usepackage{inconsolata}

\usepackage{graphicx}
\usepackage{cleveref}
\usepackage{kotex}
\usepackage{placeins} 
\usepackage{booktabs}

\title{Theme-Explanation Structure for Table Summarization using \\ Large Language Models: A Case Study on Korean Tabular Data}

\author{
TaeYoon Kwack$^{1 }$\Thanks{Both authors contributed equally to this work.} \quad Jisoo Kim$^{1 *}$\quad Ki Yong Jung$^{1}$\quad DongGeon Lee$^{2}$\quad Heesun Park$^{1 }$\thanks{Corresponding author.} \\
$^{1}$Sungkyunkwan University 
\quad 
$^{2}$Pohang University of Science and Technology\\
\texttt{\{njj05043, clrdln, wjdrldyd0213\}@g.skku.edu} \\
\texttt{donggeonlee@postech.ac.kr} \quad \texttt{hspark20@skku.edu}\\
\\
}

\begin{document}
\maketitle
\begin{abstract}
Tables are a primary medium for conveying critical information in administrative domains, yet their complexity hinders utilization by Large Language Models (LLMs). This paper introduces the \textbf{T}heme-E\textbf{x}planation Structure-based Table Summarization (\textbf{Tabular-TX}) pipeline, a novel approach designed to generate highly interpretable summaries from tabular data, with a specific focus on Korean administrative documents. Current table summarization methods often neglect the crucial aspect of human-friendly output. Tabular-TX addresses this by first employing a multi-step reasoning process to ensure deep table comprehension by LLMs, followed by a journalist persona prompting strategy for clear sentence generation. Crucially, it then structures the output into a Theme Part (an adverbial phrase) and an Explanation Part (a predicative clause), significantly enhancing readability. Our approach leverages in-context learning, obviating the need for extensive fine-tuning and associated labeled data or computational resources. Experimental results show that Tabular-TX effectively processes complex table structures and metadata, offering a robust and efficient solution for generating human-centric table summaries, especially in low-resource scenarios.

\end{abstract}

\section{Introduction}

Tables are essential for presenting core information, especially within the administrative domain, where critical data is frequently structured in tabular formats \cite{10.1007/978-3-031-60615-1_7}. The ability of Large Language Models (LLMs) to accurately summarize and elucidate the contents of these tables is becoming increasingly significant for data utilization. An important aspect of effective table summarization is the generation of human-understandable output. This necessitates not only the LLM's profound comprehension of the input table, but also the crafting of summaries that are both intuitive and concise, delivering key information without ambiguity.

Despite the critical need for human-centric summaries, recent research in table-to-text generation has often overlooked this aspect. 
Many existing approaches prioritize other metrics or model architectures, without sufficiently addressing how the generated text will be perceived and understood \cite{Liu2023RethinkingTD, Zhang2023TableLlamaTO}. 
Consequently, while the output can be factually correct, they lack the clarity, conciseness, or intuitive structure that facilitates effortless human comprehension, particularly when dealing with specialized data such as Korean administrative tables \cite{korean}.

To address this gap and emphasize the generation of human-friendly summaries, we propose the \textbf{T}heme-E\textbf{x}planation Structure-based Table Summarization (\textbf{Tabular-TX}) pipeline. Our approach is meticulously designed to guide LLMs towards producing summaries that are not only accurate but also exceptionally interpretable. First, to ensure a deep understanding of the input table, we decompose the LLM's reasoning process into a multi-step procedure, where each step focuses on a specific inferential task, thereby simplifying the complex table interpretation process. Second, we employ a ``Journalist Persona'' prompting strategy to encourage the generation of clear, objective, and well-phrased sentences. Finally, and most critically, we transform these generated insights into a highly structured Theme-Explanation (TX) format, where each summary segment consists of a thematic adverbial phrase followed by a predicative explanatory clause, enhancing readability and directness.

\begin{figure*}[!t]
    \centering
    \includegraphics[width=\textwidth]{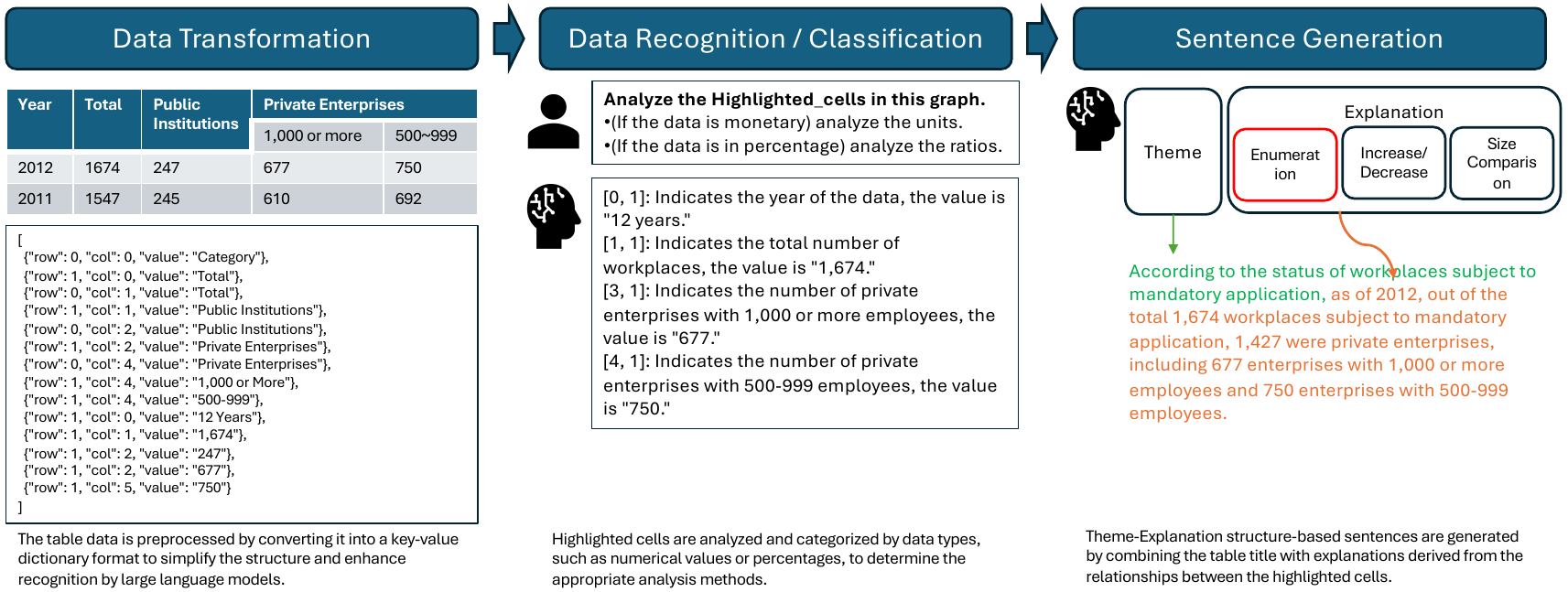}
    \caption{An overall pipeline of Theme-Explanation Structure-based Table Summarization (Tabular-TX).}
    \label{overview}
\end{figure*}

The Tabular-TX pipeline offers significant advantages, particularly in resource-constrained environments where extensive fine-tuning is unfeasible. 
Our primary contributions are threefold:
\begin{itemize}
  \item We introduce a novel pipeline that leverages multi-step reasoning and a journalist persona to generate high-quality textual explanations from complex tabular data.
  \item We propose the Theme-Explanation (TX) sentence structure, a new format for table summaries designed to maximize human interpretability. 
  \item Through empirical evaluation, we demonstrate that our In-Context Learning (ICL)-based approach enables LLMs to achieve strong performance in table data processing and summarization without the need for task-specific fine-tuning.
\end{itemize}

\section{Related Work}

\paragraph{Table-to-Text Generation}
To enable complex reasoning over tabular data, \citet{wang2024chainoftable} proposed the Chain-of-Table framework as an extension of the text-based Chain-of-Thought method \cite{cot}. This approach simplifies inference by reordering, extracting, and filtering table data, ultimately integrating relevant information into a structured table format. While this method excels in structured table processing and mathematical reasoning, it has limitations in generating interpretations for sections that require metadata or background knowledge.

TableLlama \cite{Zhang2023TableLlamaTO} aims to generalize table-based models beyond task-specific constraints by fine-tuning 14 datasets across 11 tasks, including Highlighted Cells question-answering. The model achieved performance comparable to or surpassing task-specific models and even outperformed GPT-4 \cite{GPT-4} on unseen tasks. However, despite its effectiveness, the model suffers from high computational costs.

Among table-based interpretation benchmarks, FeTaQA \cite{FeTaQA} serves as a key reference dataset. While Chain-of-Table and TableLlama utilize in-context learning \cite{few-shot-learners} and fine-tuning, respectively, they struggle to incorporate metadata into their interpretations effectively.

\section{Theme-Explanation Structure}

Unlike conventional approaches that treat table summaries as isolated text generation tasks, our method ensures structural consistency by explicitly organizing content into a \textbf{Theme Part} and an \textbf{Explanation Part}.

\subsection{Theme Part}

The Theme Part serves as a crucial contextual anchor, ensuring that numerical or categorical values in the table are interpreted correctly. It is structured as an adverbial phrase, combining the noun phrase of the table title (\texttt{table\_title}) with a citation or basis expression\footnote{Comparison of summarization sentence with and without theme part is illustrated at Figure \ref{title}.}.
 This structure is essential because the table title provides the sole comprehensive context in table summaries. Unlike general text summaries, table cells alone are insufficient to provide meaningful context, making the resulting sentence ambiguous without additional background information.

\subsection{Explanation Part}
Following the Theme Part, the Explanation Part delivers a structured analysis of the highlighted cells, forming the core content of the summary. 
Depending on the data type, this section uses a specific analytical technique, such as enumeration, magnitude comparison, or trend analysis.
The choice of method is determined based on the comparability of the highlighted cells, ensuring that the summary provides meaningful insights rather than just raw cell values.

\section{Tabular-TX Pipeline}
Generating summaries with the theme-explanation structure in Tabular-TX involves multiple processing steps to transform tabular data into structured natural language summaries.\footnote{Within Tabular-TX, data is preprocessed before using LLMs. Further details on data preprocessing are discussed in Appendix \ref{sec:appendix_data_preprocessing}.}

\subsection{Chain-of-Thought (CoT)}

After data transformation, the actual sentence generation process begins using Chain-of-Thought (CoT) reasoning. Large Language Models (LLMs) generally face performance degradation when handling tabular data summarization, as this task simultaneously requires multiple capabilities such as table recognition, mathematical reasoning, and commonsense inference. This issue, known as the Compositional Deficiency problem \cite{zhao2024compositionaldeficiency}, occurs because individual data points tend to be analyzed separately without adequately integrating their relationships into a holistic interpretation. CoT mitigates this by systematically guiding the model to tackle one reasoning step at a time, thereby improving interpretative accuracy and contextual completeness.

Specifically, CoT first classifies the types of highlighted cells, distinguishing among monetary values, percentages, categorical data, and textual explanations. This classification step prevents potential errors, such as misinterpreting percentages as plain numbers. Then, depending on the classified data type, the most appropriate analytical method---such as enumeration for listing individual items, magnitude comparison for numerical rankings, or trend analysis for temporal changes---is selected and applied. For instance, monetary values are converted into consistent units, and percentages are appropriately formatted for clarity.

In the context of Korean administrative table data, these challenges are further complicated by the language's implicit nature, the potential gap between administrative and everyday terminology, and morphological complexities (e.g., absent subjects or ambiguous particle usage). CoT systematically decomposes these linguistic hurdles by (1) classifying specialized terms, (2) normalizing numeric expressions in line with Korean usage conventions, and (3) incrementally integrating contextual cues, such as clarifying administrative vocabulary or disambiguating omitted referents. Through this stepwise process, the model avoids misinterpretations caused by either unfamiliar terms or implicit structures, ultimately generating summaries that better align with Korean textual norms.

By addressing each reasoning subtask explicitly and sequentially, CoT ensures the final table summary captures data relationships clearly and coherently, resulting in an accurate and contextually meaningful summary.

\subsection{Journalist Persona for Structured Generation}
 In last step of pipeline we assign a journalist persona to the LLM to generate Theme-Explanation structured summaries. This persona is particularly effective because table summaries share key characteristics with straight news articles, which prioritize conciseness, objectivity, and fact-based clarity. Rather than generating overly detailed or speculative content, the model produces well-structured and neutral summaries that adhere to journalistic reporting conventions when guided by this persona.

Figure~\ref{persona} demonstrates the impact of the journalist persona on table summarization. With a generic prompt, the model generates an ambiguous summary that captures core information but lacks contextual clarity and coherence. In contrast, applying the journalist persona produces a structured and contextually enriched summary. This improvement occurs because the journalist persona explicitly guides the model to state information sources, clearly define numerical constraints, and incorporate contextual details, closely resembling news article formats.

\begin{table*}[!h]
\centering
\footnotesize

\resizebox{\textwidth}{!}{%

\begin{tabular}{@{}l|ccc|c@{}}
\toprule
Model                                            & \multicolumn{1}{l}{ROUGE-1} & \multicolumn{1}{l}{ROUGE-L} & \multicolumn{1}{l|}{BLEU} & \multicolumn{1}{l}{Average} \\ \midrule
KoBART - Fine-tuned                      & 0.37                        & 0.28                        & 0.35                      & 0.33                        \\ \midrule
EXAONE 3.0 7.8B - ICL                            & 0.21                        & 0.14                        & 0.01                      & 0.12                        \\
EXAONE 3.0 7.8B - LoRA                           & 0.27                        & 0.21                        & 0.05                      & 0.17                        \\
\textbf{EXAONE 3.0 7.8B - Tabular-TX}            & \textbf{0.51}               & \textbf{0.39}               & \textbf{0.44}             & \textbf{0.45}               \\ \midrule
llama-3-Korean-Bllossom-8B - ICL                 & 0.33                        & 0.25                        & 0.27                      & 0.28                        \\
\textbf{llama-3-Korean-Bllossom-8B - Tabular-TX} & \textbf{0.48}               & \textbf{0.37}               & \textbf{0.42}             & \textbf{0.43}               \\ \bottomrule
\end{tabular}

}
\caption{Evaluation results on the Korean Korean table interpretation benchmark for each model.}
\label{results}
\end{table*}

\section{Experiment}

\subsection{Experimental Setup}

\paragraph{Dataset}
For training and evaluation, we utilized the Korean table interpretation benchmark \cite{korean}, which focuses on summarizing highlighted table segments into coherent sentences. An example of the dataset is shown in Figure~\ref{sample}. The dataset consists of 7,170 training tables, 876 validation tables, and 876 test tables. Each data point contains metadata such as the document title, table title, publication date, publishing organization, table source URL, highlighted cell information, table data, and a reference summary sentence describing the highlighted portions' key contents.

\paragraph{Evaluation Metrics}
We employ ROUGE-1, ROUGE-L, and BLEU to evaluate the performance of table segment interpretation. These metrics assess how effectively the summaries convey the key content of the table while achieving high semantic quality.

\paragraph{Models}
To evaluate the effectiveness of the Tabular-TX pipeline, we utilize EXAONE 3.0 7.8B \cite{lg-exaone} and llama-3-Korean-Bllossom-8B\footnote{\url{https://huggingface.co/MLP-KTLim/llama-3-Korean-Bllossom-8B}} models as base models. EXAONE 3.0 7.8B, a successor to EXAONE-LM-v1.0, has demonstrated state-of-the-art performance in Korean TableQA, ranking first on the KorWikiTableQuestions \cite{KorWikiTableQuestions}. Similarly, llama-3-Korean-Bllossom-8B is the top-performing sub-10B model in a Korean multi-domain reasoning benchmark.
We compare the performance of these models with and without Tabular-TX, assessing whether structured generation enhances performance in table summarization. Additionally, we analyze whether Tabular-TX reduces reliance on extensive fine-tuning while maintaining high-quality summaries.

\subsubsection{Additional Adaptation Approaches}

\paragraph{In-Context Learning (ICL)} We begin by applying ICL to each model, providing a few table-summarization examples without explicit fine-tuning. This approach tests how effectively the model can generate coherent sentences for highlighted table cells based solely on a small set of demonstrations.

\paragraph{Low-Rank Adaptation (LoRA)} Next, we assess the computational efficiency and performance of the Tabular-TX pipeline by introducing LoRA \cite{LoRA}.  We trained the EXAONE 3.0 7.8B by applying LoRA to see if we could maintain high-quality table summaries with fewer resources.

\paragraph{Full Model Fine-Tuning} Finally, we evaluate the KoBART (Korean BART)\footnote{\url{https://huggingface.co/gogamza/kobart-base-v2}} model under a full model fine-tuning setup to determine whether a smaller-scale language model can achieve comparable performance when all its parameters are trained on the Korean table interpretation benchmark.

\subsection{Experimental Results}
Table~\ref{results} presents the performance of various models evaluated using ROUGE-1, ROUGE-L, BLEU, and their average scores. The KoBART recorded an average score of 0.33 after fine-tuning. In contrast, EXAONE 3.0 7.8B achieved 0.12 with the ICL method, 0.17 after fine-tuning, and 0.45 when combined with the Tabular-TX method. Similarly, llama-3-Korean-Bllossom-8B, which was also tested with Tabular-TX, showed a notable improvement, reaching an average score of 0.43. These results demonstrate that Tabular-TX consistently outperforms alternative methods, achieving the highest overall performance across different model configurations.

The performance gap between EXAONE 3.0 7.8B and KoBART, despite both being fine-tuned on the same dataset, can be explained through the multiplicative joint scaling law \cite{zhang2024scalingmeetsllmfinetuning}. This principle suggests that when the dataset size is insufficient relative to the model size, the performance gains from fine-tuning remain limited. Since KoBART has 124M parameters, while EXAONE 3.0 is approximately 63 times larger, the dataset required to achieve a comparable performance boost must be proportionally scaled up by a factor of 63. The inability to meet this scaling requirement explains why the performance of the fine-tuned KoBART plateaued, while EXAONE 3.0 7.8B demonstrated more significant gains with the same dataset.

This study confirms that the proposed Tabular-TX method enhances table data analysis performance without fine-tuning. Notably, Tabular-TX outperforms traditional fine-tuned models despite relying on significantly smaller datasets, demonstrating its efficacy in resource-constrained learning environments. Moreover, Tabular-TX achieved approximately four times higher average performance compared to standard ICL methods, further reinforcing its role as a scalable and efficient alternative for structured table summarization tasks.

\section{Conclusion}

This study introduced the Theme-Explanation Table Summarization (Tabular-TX) pipeline, a novel approach to improve table summarization tasks with low-resource requirements. Moreover, the proposed pipeline effectively overcame unique challenges in Korean administrative table processing.

Experimental results signify that Tabular-TX enhances table summarization performance. This study contributed to the summarization of complex table data by introducing a novel sentence generation method based on the theme-explanation structure. Furthermore, Tabular-TX achieved excellent performance without fine-tuning, by incorporating ICL. This indicates its potential as a significant contribution to table data analysis, even in resource-constrained environments, without requiring direct model training.

\section*{Limitations}

We acknowledge a few limitations in this study.
First, Tabular-TX was only evaluated on EXAONE 3.0 7.8B and llama-3-Korean-Bllossom-8B, leaving the question of its effectiveness across a broader range of LLMs open.
Second, this study primarily focused on Korean administrative table data, and further research should investigate whether the Theme-Explanation structure is equally effective for diverse tabular data formats in other languages or specialized domains.
Finally, Tabular-TX currently relies on predefined structural components (Theme and Explanation parts) to enforce interpretability. Future work should explore more dynamic approaches that allow for adaptive sentence structuring based on different types of tables, potentially improving performance across varied tabular structures.

\bibliography{custom}
\clearpage
\onecolumn
\appendix
\label{sec:appendix}

\begin{figure}[!ht]
    \centering
    \fbox{\includegraphics[width=\columnwidth]{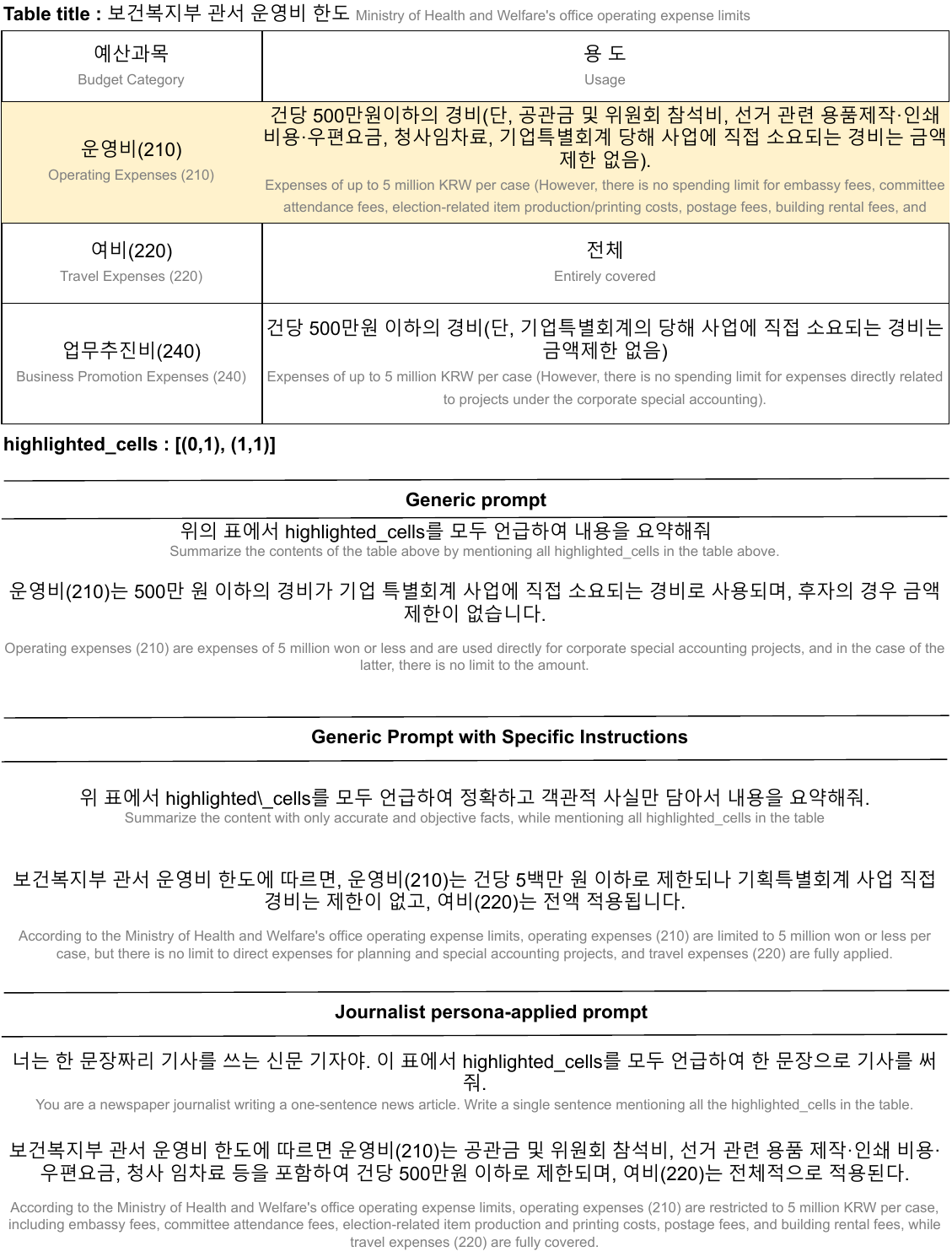}}
    \caption{An example of how a persona-applied prompt generates more accurate summaries. (Gray text indicates the English translation.)}
    \label{persona}
\end{figure}

\clearpage

\section{Korean table interpretation benchmark}
\label{sec:appendix_example}
\begin{figure}[!ht]
    \centering
    \fbox{\includegraphics[width=\columnwidth]{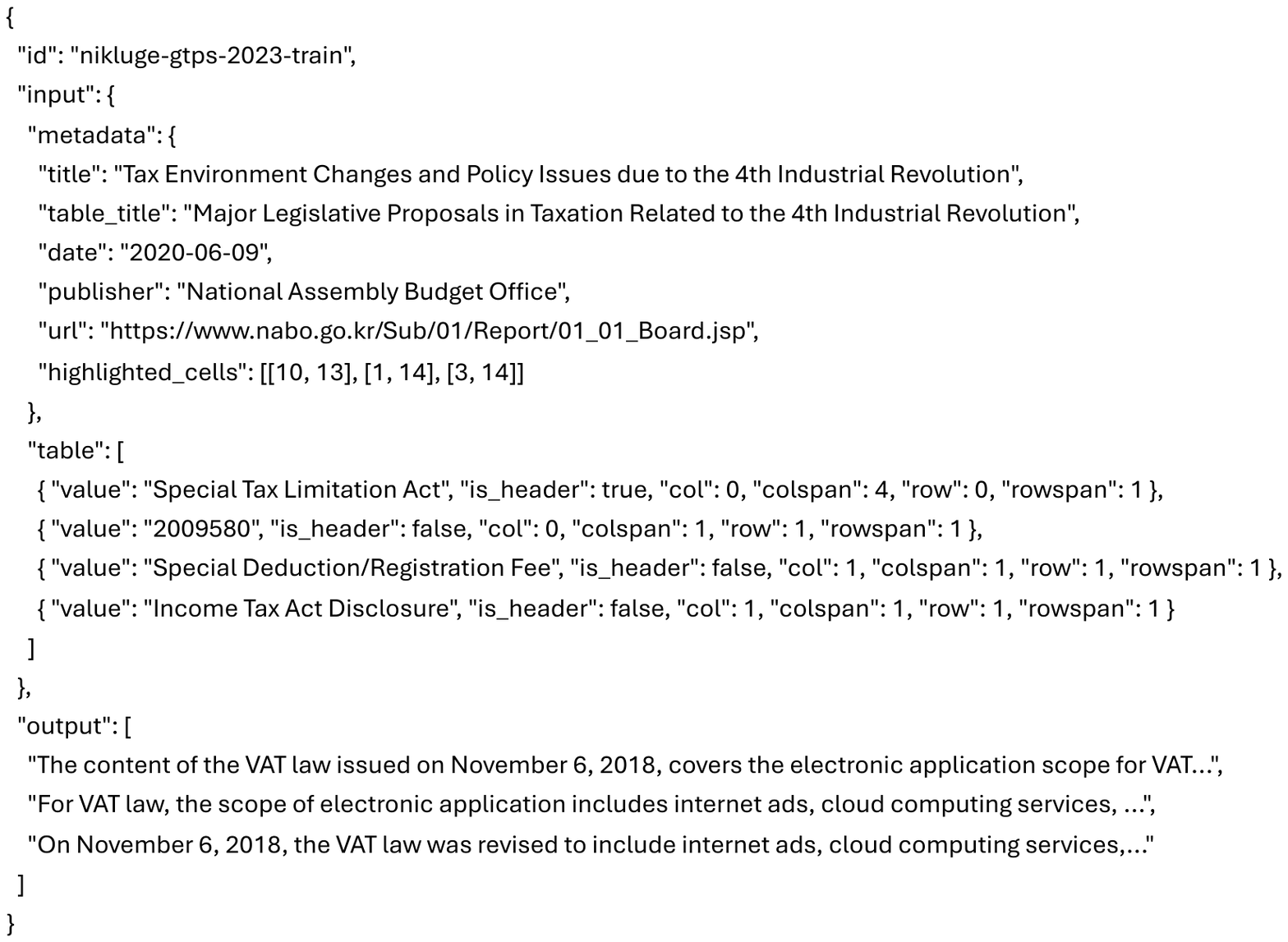}}
    \caption{An example from the corpus for evaluating interpretation generation of table segments (originally in Korean, translated into English) \cite{korean}.}
    \label{sample}
\end{figure}

We leverage the Korean table interpretation benchmark provided by the National Institute of Korean Language \cite{korean}.

As shown in Figure~\ref{sample}, an objective of the dataset is to summarize the highlighted cells, which are labeled in the metadata as \texttt{highlighted\_cells}, into a single coherent sentence.

\clearpage

\section{Examples of the Theme Part \& Explanation Part}

\begin{figure}[!ht]
    \centering
    \fbox{\includegraphics[width=\columnwidth]{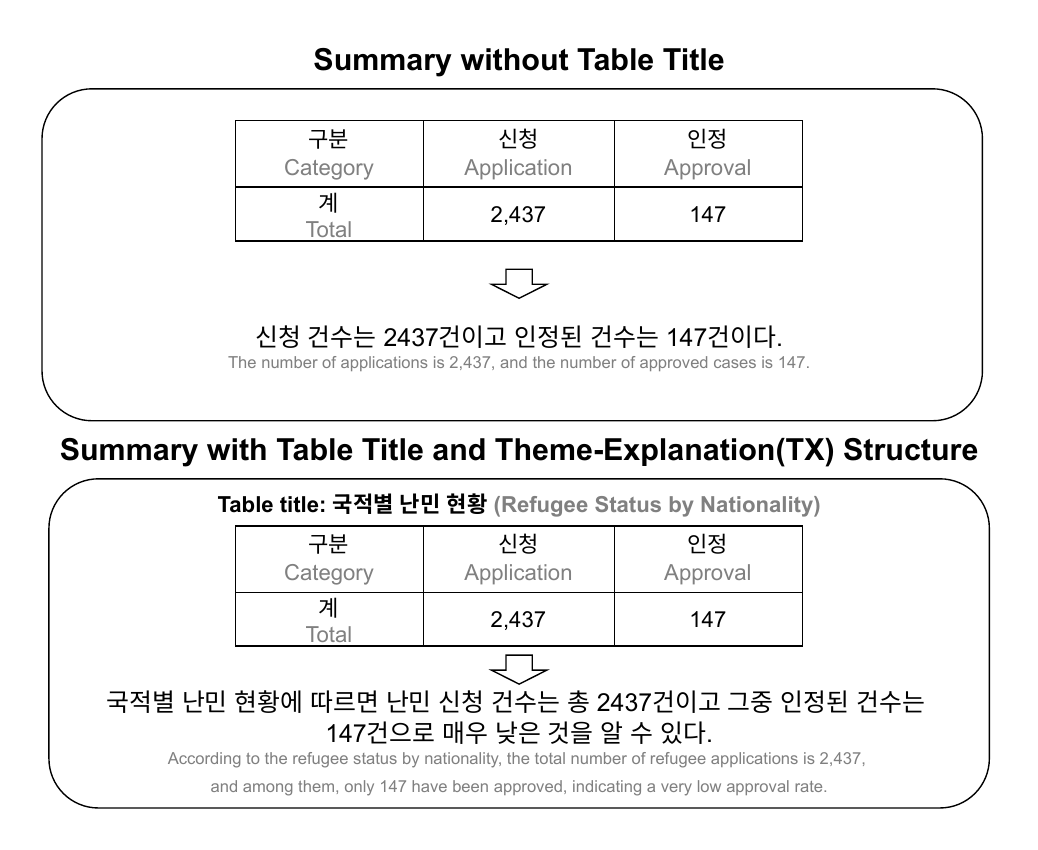}}
    \caption{A sentence including the table title conveys the context more accurately. (Gray text indicates the English translation.)}
    \label{title}
\end{figure}

\subsection{Theme Part}

For example, in the sentence: \textit{``According to the refugee status by nationality, the total number of refugee applications is 2,437,
and among them, only 147 have been approved, indicating a very low approval rate.''} Here, the Theme Part is: \textit{``According to the refugee status by nationality,''} This phrase, introduced with the citation expression \textit{``According to''}, provides essential context for the numerical values that follow. Without this structured introduction, the reader may struggle to understand the significance of the numbers. Figure~\ref{title} illustrates how omitting the Theme Part results in an unclear or misleading summary.

\subsection{Explanation Part}

For instance, in the previous example, the Explanation Part is: \textit{``the net fiscal cost increased by 9.435 trillion KRW from the previous year, reaching a total of 61.301 trillion KRW.''} Here, the Explanation Part is derived by comparing the numerical changes between the two cells. Here, a trend analysis is applied to highlight the increase in fiscal cost.

\section{Implementation Details}
\subsection{Data Preprocessing}
\label{sec:appendix_data_preprocessing}

The first step is preprocessing the table to simplify its structure for better LLM comprehension. 
Since LLMs primarily operate on sequential text representations, directly processing raw tabular formats can lead to misinterpretation of hierarchical relationships within the data. To address this, we convert table data into a key-value pair dictionary format, which is commonly used in natural language processing tasks. This transformation significantly enhances LLMs' ability to recognize table semantics, improving summarization accuracy \cite{Parsing}.

\begin{figure}[!ht]
    \centering
    \fbox{\includegraphics[width=\columnwidth]{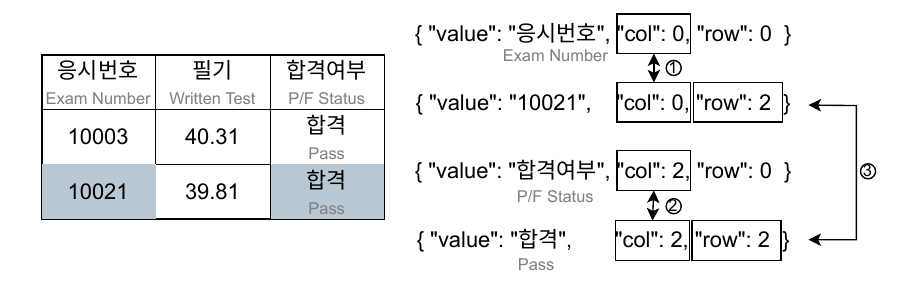}}
    \caption{An example of inferring relationships between data sharing the same row or column. Through inference in ①, it is deduced that `10021' represents the `Exam number.' In ②, the meaning of `pass' is inferred. In ③, it is deduced that the exam with the `Exam number' `10021' has `passed.' (Gray text indicates the English translation.)}
    \label{proc}
\end{figure}

\begin{figure}[!ht]
    \centering
    \fbox{\includegraphics[width=\columnwidth]{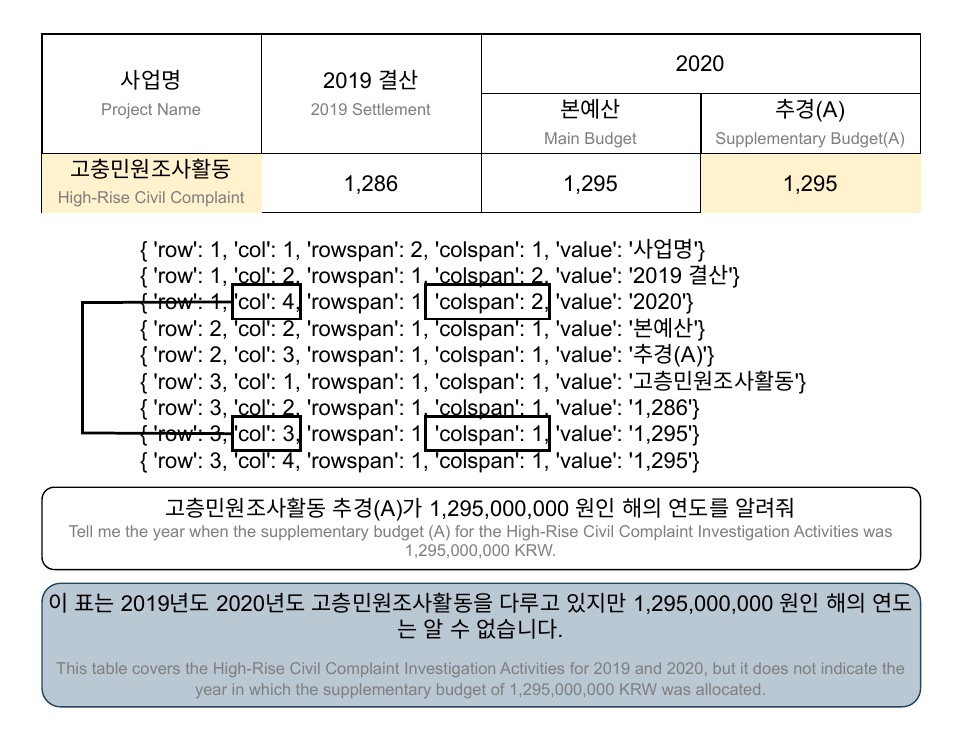}}
    \caption{An example of how merged cells hinder table recognition. (Gray text indicates the English translation.)}
    \label{migrated_cell}
\end{figure}

Then, we process merged cells to clarify the table structure. Merged cells span multiple rows or columns and are defined by `rowspan' and `colspan.' As shown in Figure~\ref{proc}, LLMs infer relationships between data through row or column alignment. However, incorrect handling of merged cell ranges can lead to misinterpretation.
For example, in Figure~\ref{migrated_cell}, the cell labeled ``2020'' should cover columns 3 and 4, but it appears only in column 3. To resolve this, merged cells are replicated across their ranges, allowing LLMs to recognize cell dependencies and hierarchical structures correctly.

Finally, the transformed dictionary list retains only the highlighted and related cells, where ``related cells'' refer to all header cells sharing the same row/column as the highlighted cells. This process reduces data complexity and enhances LLMs' recognition of table structures.

\clearpage

\subsection{Prompt Details}

\begin{figure}[!ht]
    \centering
    \fbox{\includegraphics[width=\columnwidth]{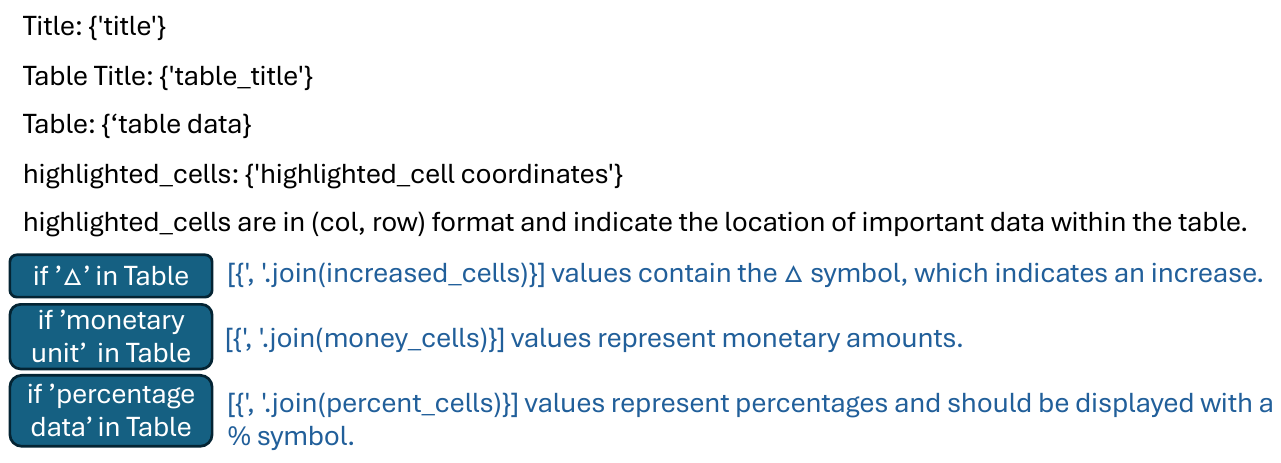}}
    \caption{Summarizing key data points from the table in a single sentence for a news article. (originally in Korean, translated into English)}
    \label{prompt_1}
\end{figure}

\begin{figure}[!ht]
    \centering
    \fbox{\includegraphics[width=\columnwidth]{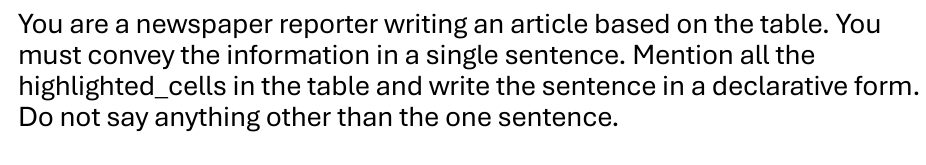}}
    \caption{Writing a one-sentence summary of a table by embodying news reporter persona. (originally in Korean, translated into English)}
    \label{prompt_2}
\end{figure}

Tabular-TX generates table summaries in two steps.  
First, the \textbf{Data Recognition / Classification} step identifies key data from highlighted cells (Figure~\ref{prompt_1}).  
Second, the \textbf{Sentence Generation} step forms a summary in the Theme-Explanation format (Figure~\ref{prompt_2}).


\end{document}